\documentclass{article}

\usepackage{arxiv}

\date{}

\usepackage{graphicx}
\usepackage{booktabs} 

\usepackage{url}
\usepackage[utf8]{inputenc} 
\usepackage[T1]{fontenc}    
\usepackage{url}            
\usepackage{booktabs}       
\usepackage{amsfonts}       
\usepackage{microtype}      
\usepackage{amsmath}
\usepackage{subcaption}
\usepackage{graphicx}
\usepackage{amssymb}
\usepackage{xcolor}
\usepackage{natbib}

\usepackage[frozencache,cachedir=.]{minted}

\usepackage{filecontents}
\usepackage{pgfplots}

\makeatletter
\let\c@figure\c@table
\let\ftype@figure\ftype@table
\let\ext@figure\ext@table

\makeatother 

\usepackage{hyperref}       

\title{Self-attention Does Not Need $O(n^2)$ Memory}

\author{Markus N. Rabe and Charles Staats \\
Google Research\\
\texttt{\{mrabe,cstaats\}@google.com}
}

\begin{document}

\maketitle

\begin{abstract}
We present a very simple algorithm for attention that requires $O(1)$ memory with respect to sequence length and an extension to self-attention that requires $O(\log n)$ memory.
This is in contrast with the frequently stated belief that self-attention requires $O(n^2)$ memory.
While the time complexity is still $O(n^2)$, device memory rather than compute capability is often the limiting factor on modern accelerators.
Thus, reducing the memory requirements of attention allows processing of longer sequences than might otherwise be feasible.
We provide a practical implementation for accelerators that requires $O(\sqrt{n})$ memory, is numerically stable, and is within a few percent of the runtime of the standard implementation of attention.
We also demonstrate how to differentiate the function while remaining memory-efficient.
For sequence length 16384, the memory overhead of self-attention is reduced by 59X for inference and by 32X for differentiation.
\end{abstract}

\section{Introduction}

Attention~\citep{BahdanauCB2015attention} is widely used in modern neural architectures.
In particular, it is the heart of the Transformer architecture~\citep{vaswani2017attention}, which has revolutionized Natural Language Processing~\citep{devlin2019bert}, and found wide-spread adoption across several research areas since then.

Given a query $q\in\mathbb{R}^d$ and lists of keys and values $k_1,\dots,k_n$ and $v_1,\dots,v_n\in\mathbb{R}^{d}$ of length $n$, \emph{attention} is defined as follows:
\[
s_i = \mathrm{dot}(q, k_i),\qquad s'_i = \frac{e^{s_i}}{\sum_j e^{s_j}}, \qquad \mathrm{attention}(q,k,v)=\sum_i v_i s'_i.
\]
The result of the attention operation for a single query, is hence a weighted sum of the value vectors, where the weights are the softmax of the dot products of the query and the keys.

The straight-forward implementation of the attention operation above requires us to first compute and remember $s_i$ for all $i$, leading to a $O(n)$ time and memory complexity for each query.
Transformers use \emph{self-attention}, which issues a separate query for each position in the sequence, so the overall time and space complexity is $O(n^2)$.

In many works the quadratic time and space complexity of self-attention has been used as the motivation for the investigation of variants of the original attention mechanism and architectures with more favorable complexity classes~\citep{kitaev2020reformer,RoySVG2021Routing,ZaheerGDAAOPRWY2020BigBird,ChoromanskiLDSGSHDMKBCW2020Performer,wang2020linformer,ren2021combiner,ChildGRS2019SparseTransformers,Tay0ASBPRYRM2021LongRangeArena,wang2020linformer,MaKWZMMZ2021Luna,ShenZZY2021EfficientAttention,qiu2020blockwise}.
Modern accelerator hardware, such as GPUs and TPUs, are often memory constrained for applications in deep learning, while compute is relatively cheap.
So the space complexity of transformers is a particular concern, c.f.~\citet{kitaev2020reformer,RoySVG2021Routing,ZaheerGDAAOPRWY2020BigBird}.

In this work, we present new algorithms for attention and self-attention that require only constant memory and logarithmic memory, respectively.
The basic algorithm is very simple; but it requires a trick to make it numerically feasible (see Section~\ref{sec:numerics}).
We also present an implementation in JAX~\citep{jax2018github}, which runs efficiently on TPUs, and requires $O(\sqrt{n})$ memory for self-attention (see Section~\ref{sec:implementation}).

Unlike other works that aim to reduce the memory complexity of attention, the memory-efficient algorithm for attention that we suggest is not an approximation, but computes the same function.
We can hence use the memory-efficient algorithm as a drop-in replacement for other attention implementations to save memory.
This may allow us to reconsider architecture choices, or scale to new datasets that require longer, dense attention.
%
However, our algorithm still requires $O(n^2)$ time complexity for self-attention and $O(n)$ time complexity for single-query attention, and the various efficient, long-context attention mechanisms remain an interesting alternative to (dense) attention.


\section{Algorithm}
\label{sec:algorithm}

First, we present the algorithm for the attention operation with a single query and extend the algorithm to self-attention at the end of this Section.
We observe that the division by $\sum_j e^{s_j}$ can be moved to the very end of the attention operation using the distributive law:
\begin{equation}\label{eq:lazySoftmax}
s_i = \mathrm{dot}(q, k_i),\qquad s'_i = e^{s_i}, \qquad \mathrm{attention}(q,k,v)=\frac{\sum_i v_i s'_i}{\sum_j s'_j}.
\end{equation}

After publishing our initial draft, we were made aware that \eqref{eq:lazySoftmax} is a rediscovery of the ``lazy softmax" method of \citet[equation 4]{HanhwiJJJJ2019MNNFast}. Unfortunately their
paper went in a different direction and did not discuss the memory complexity implications and
other innovations we present in the remainder of this paper. For more details see Section \ref{section:relatedWork}.

This can be computed with constant memory:
The memory overhead of this algorithm consists of a vector $v^*\in\mathbb{R}^d$ and a scalar $s^*\in\mathbb{R}$, both initialized with 0.
Given the query $q$, keys $k_1, \dots, k_n$ and values $v_1,\dots,v_n$, we process the keys and values in sequence.
Given a key value pair $k_i$, $v_i$, we compute $s_i=\mathrm{dot}(q, k_i)$ and update $v^* \leftarrow v^* + v_i e^{s_i}$ and $s^* \leftarrow s^* + e^{s_i}$.
After processing all keys and values, we divide $\frac{v^*}{s^*}$ to get the final result.

The analysis of space complexity assumes that inputs are given in a particular order: we first read the query, and then a list of \emph{pairs} of keys and values.
If the inputs are provided in a different order, we have to additionally store an index into the sequence, requiring $O(\log n)$ memory instead.

To extend this algorithm to \emph{self-}attention, we compute the results to all queries sequentially.
This requires just one additional index into the list of queries, giving rise to the $O(\log n)$ memory complexity.
Note that the operation produces outputs that are linear in the size of the number of queries, i.e., $O(n)$, which is not counted towards the space complexity.

\section{Numerical Stability}
\label{sec:numerics}

The formulation of standard attention that we presented in the Introduction, as well as our memory-efficient algorithm, are not numerically stable when using floating point arithmetic, because the softmax exponentiates the scores.
For scores $\geq89$ the exponentiation results in \texttt{inf} (for \texttt{bfloat16} and \texttt{float32}), which will be carried through to the final result of the attention operation.
In practice, the softmax is implemented by subtracting the maximum score from all scores.
This does not change the result of the softmax, but avoids this numerical problem.

Our incremental computation of the sum of exponentiated scores (and the values times the scores) does not immediately allow for the same trick, as the maximum may depend on the last score in the sequence.
But the subtraction cannot be delayed either, since the scores must be exponentiated before they can be added to the cumulative sum.

To resolve this problem, we introduce an additional scalar, which keeps track
of the maximum score that the incremental algorithm has seen so far, and we renormalize the sums of exponentiated values as needed:
We initialize the vector $v^*\in\mathbb{R}^d$ and scalar $s^*\in\mathbb{R}$ with 0, and $m^*$ with $-\mathrm{inf}$.
As before, given a key value pair $k_i$, $v_i$, we compute $s_i=\mathrm{dot}(q, k_i)$, but then the algorithm differs slightly from Section~\ref{sec:algorithm}.
We first compute $m_i=\max(m^*, s_i)$ and update $v^* \leftarrow v^* e^{m^*-m_i} + v_ie^{s_i-m_i}$ and $s^* \leftarrow s^*e^{m^*-m_i} + e^{s_i-m_i}$ and $m^*\leftarrow m_i$.
After processing all keys and queries, we divide $\frac{v^*}{s^*}$ to get the final result.

\section{An Implementation For TPUs}
\label{sec:implementation}

\begin{figure}
\vspace{-5mm}
\begin{minted}[xleftmargin=1em,linenos]{python}
import functools, jax, math
from jax import numpy as jnp

def _query_chunk_attention(query, key, value, precision, key_chunk_size=4096):
  """Multi-head dot product attention with a limited number of queries."""
  num_kv, num_heads, k_features = key.shape
  v_features = value.shape[-1]
  key_chunk_size = min(key_chunk_size, num_kv)
  query = query / jnp.sqrt(k_features)

  @functools.partial(jax.checkpoint, prevent_cse=False)
  def summarize_chunk(query, key, value):
    attn_weights = jnp.einsum('qhd,khd->qhk', query, key, precision=precision)
    max_score = jnp.max(attn_weights, axis=-1, keepdims=True)
    max_score = jax.lax.stop_gradient(max_score)
    exp_weights = jnp.exp(attn_weights - max_score)
    exp_values = jnp.einsum('vhf,qhv->qhf', value, exp_weights, precision=precision)
    return (exp_values, exp_weights.sum(axis=-1),
            max_score.reshape((query.shape[0], num_heads)))

  def chunk_scanner(chunk_idx):
    key_chunk = jax.lax.dynamic_slice(
        key, (chunk_idx, 0, 0),
        slice_sizes=(key_chunk_size, num_heads, k_features))
    value_chunk = jax.lax.dynamic_slice(
        value, (chunk_idx, 0, 0),
        slice_sizes=(key_chunk_size, num_heads, v_features))
    return summarize_chunk(query, key_chunk, value_chunk)

  chunk_values, chunk_weights, chunk_max = jax.lax.map(
      chunk_scanner, xs=jnp.arange(0, num_kv, key_chunk_size))

  global_max = jnp.max(chunk_max, axis=0, keepdims=True)
  max_diffs = jnp.exp(chunk_max - global_max)
  chunk_values *= jnp.expand_dims(max_diffs, axis=-1)
  chunk_weights *= max_diffs

  all_values = chunk_values.sum(axis=0)
  all_weights = jnp.expand_dims(chunk_weights, -1).sum(axis=0)
  return all_values / all_weights

def attention(query, key, value, precision=jax.lax.Precision.HIGHEST,
              query_chunk_size=1024):
  """Memory-efficient multi-head dot product attention."""
  num_q, num_heads, q_features = query.shape

  def chunk_scanner(chunk_idx, _):
    query_chunk = lax.dynamic_slice(
        query, (chunk_idx, 0, 0),
        slice_sizes=(min(query_chunk_size, num_q), num_heads, q_features))
    return (chunk_idx + query_chunk_size,
            _query_chunk_attention(query_chunk, key, value, precision=precision))

  _, res = jax.lax.scan(
      chunk_scanner, init=0, xs=None, length=math.ceil(num_q / query_chunk_size))
  return res.reshape(num_q, num_heads, value.shape[-1])
\end{minted}
\caption{Implementation of memory-efficient attention suited for TPUs.}
\label{fig:my_label}
\end{figure}

In this section, we provide a version of the algorithm above that exploits the massive parallelism of modern hardware, such as GPUs or TPUs.
The naive algorithm above is is not trivial to parallelize for a compiler, as the incremental sum introduces a dependency across all keys and values.

We present the entire implementation, including the support for multiple attention heads and memory-efficient differentiation in Figure~\ref{fig:my_label}.
The implementation does not optimize strictly for memory efficiency, but instead aims to strike a balance between simplicity, computational efficiency, and memory requirements.

To exploit the parallelism available in modern hardware, we split the computation into chunks at the cost of some additional memory.
In the outer loop (lines 54-55), we split the queries in to chunks of constant size, resulting in a linear number of iterations.
In each iteration of the outer loop, we call \texttt{\_query\_chunk\_attention}, which itself processes the keys and values in chunks (lines 30-31).
The chunks are processed sequentially and each chunk is summarized independently (lines 12 to 19).
Assuming a chunk size of $\sqrt{n}$ for the keys and values, we hence obtain $\sqrt{n}$ summaries, giving rise to the $O(\sqrt{n})$ memory complexity.

After the summaries are computed, they need to be rescaled (lines 33 to 36) along the lines of Section~\ref{sec:numerics}, before we return the values divided by the weights (line 40).
The result of each iteration of the outer loop is directly written to the output tensor \texttt{res} (line 54), so that no additional memory is consumed across iterations.
(A multi-stage summarization approach could achieve $O(\log n)$ but would complicate the implementation.)

While a constant chunk size for the queries and a chunk size of $\sqrt{n}$ for the keys and values is optimal for memory consumption, the runtime is also affected by the choice of chunk size in practice, which is heavily affected by the choice of hardware.
Ultimately, we have to leave this trade-off to the programmer, and expose the chunk sizes as arguments \texttt{query\_chunk\_size} and \texttt{key\_chunk\_size}.
In Figure~\ref{fig:my_label} we provide default values for the chunk sizes that lead to minimal runtime impact on TPU, while still providing significant memory savings.

\section{Empirical Analysis}
\label{sec:experiments}

In this section, we experimentally compare the memory requirements and runtime performance of the suggested algorithm compared to the implementation of attention currently provided by Flax (\citet{flax2020github}, see \texttt{flax/linen/attention.py}).
We open-sourced the code of our implementation and most of the evaluation as a colab to help others reproduce the results: \url{https://github.com/google-research/google-research/tree/master/memory_efficient_attention}.

\subsection{Inference}
\begin{table}[]
    \centering
    \begin{tabular}{l|c|c|c|c|c|c|c}
    Sequence length &$n=2^8$ & $2^{10}$ & $2^{12}$ & $2^{14}$ & $2^{16}$ & $2^{18}$ & $2^{20}$
        \\\hline\hline
        Size of inputs and outputs & 160KB & 640KB & 2.5MB & 10MB & 40MB & 160MB & 640MB \\
        Memory overhead of standard attention & 270KB & 4.0MB & 64MB & 1GB & OOM & OOM & OOM \\
        Memory overhead of memory-eff. attn. & 270KB & 4.0MB & 16MB & 17MB & 21MB & 64MB & 256MB \\
        Compute time on TPUv3 & 0.06ms & 0.11ms & 0.7ms & 11.3ms & 177ms & 2.82s & 45.2s \\
        Relative compute speed & $\pm$5\%  & $\pm$5\%  & -8$\pm$2\% & -13$\pm$2\% & - & - & - \\
    \end{tabular}
    \vspace{3mm}
    \caption{Memory and time requirements of self-attention during \textbf{inference}.}
    \label{tab:performance}
\end{table}

In Table~\ref{tab:performance} we compare the memory requirements and the compute time of the memory-efficient attention implementation and the Flax implementation of attention.
The size of inputs and outputs includes the query, key, and value tensors of dtype \texttt{bfloat16}, and the output tensor of dtype \texttt{float32}.
We measure the memory overhead as the TPUs peak memory in excess of the input and output tensors.
All computations were done on a single TPUv3 chip.
For this experiment, we only use one attention head.

Our memory-efficient implementation of attention removes the memory bottleneck of self-attention, scaling at least to a sequence length of 1M.
At this sequence length the algorithm is multiplying over 1 trillion combinations of queries and keys. The time complexity is still quadratic.

The ``relative compute speed'' of the implementations was computed as the median over 100 runs---but the numbers still fluctuated across multiple runs of the evaluation and we only provide them to demonstrate that the runtime performance is roughly similar.
Please note that this experiment analyzes the attention operation in isolation; the measured relative performance is not necessarily the same when the operations are embedded in larger architectures.
In fact, we observed a slight increase in steps/sec of about 4\% when training a small Transformer.

For all cases where the standard attention would not OOM (i.e. require $>16$GB device memory), we checked that the results of the two implementations are within $1.8 \times 10^{-7}$ for inputs drawn from a normal distribution with standard deviation $1$ (measured as the maximal absolute difference of any dimension in a self-attention over sequence length~$2^{14}$).

\subsection{Differentiation}

\begin{table}
    \centering
    \begin{tabular}{l|c|c|c|c|c|c|c}
    Sequence length &$n=2^8$ & $2^{10}$ & $2^{12}$ & $2^{14}$ & $2^{16}$ & $2^{18}$ & $2^{20}$
        \\\hline\hline
        Size of inputs and outputs & 192KB & 768KB & 2.9MB & 12MB & 47MB & 188MB & 750MB \\
        Memory overhead of standard attention & 532KB & 8.0MB & 128MB & 2.0GB & OOM & OOM & OOM \\
        Memory overhead of memory-eff. attn. & 532KB & 8.0MB & 41MB & 64MB & 257MB & 1.0GB & 4.0GB \\
        Compute time on TPUv3 & 0.1ms & 0.18ms & 1.4ms & 21ms & 336ms & 5.3s & 85s \\
        Relative compute speed & $\pm$5\% & $\pm$5\% & -30$\pm$5\% & -35$\pm$5\% & - & - & - \\
    \end{tabular}
    \vspace{3mm}
    \caption{Memory and time requirements of self-attention during \textbf{differentiation}. Note that the slowdown in compute speed is expected due to the use of checkpointing in memory-efficient attention.}
    \label{tab:differentiation}
\end{table}

During the forward pass our algorithm saves memory by summarizing parts of the attention matrix sequentially, allowing it to forget the parts of the attention matrix it has summarized already.
A naive application of differentiation would have to store all those intermediate results and our algorithm would loose its memory advantage entirely.
So we apply checkpointing~\citep{ChenXZG2016checkpointing} in line 11 to the function that summarizes the individual chunks.
The intermediate results can thus be forgotten during the forward pass and
recomputed during backpropagation.

In Table~\ref{tab:differentiation} we compare runtime and peak memory during differentiation of our implementation to standard attention.
We used the same setting as for the forward pass, but applied \texttt{jax.grad} to an arbitrarily chosen loss function (the sum of the results).
%
The relative compute speed was reduced significantly compared to standard attention. 
This is expected when using checkpointing since some values must be recomputed during backpropagation.

Note that applying checkpointing to the standard attention algorithm would not achieve these results.
The standard algorithm with checkpointing would forget the attention matrix after it is formed; our algorithm never forms the full attention matrix at all.

\subsection{Training}
\label{sec:training}

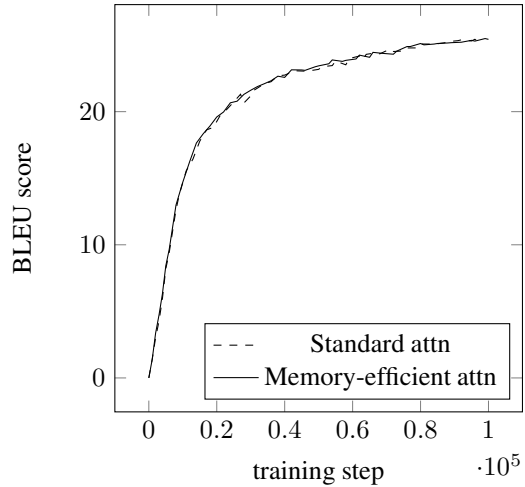
\begin{figure}
    \centering
    \begin{tikzpicture}
    \begin{axis}[ 
    width=7cm,
    height=7cm,
    no marks,
    legend pos=south east,
    xlabel={training step},
    ylabel={BLEU score},
    ]
    \addplot[dashed] table [x=Step, y=Value, col sep=comma] {standard.csv};
    \addplot[] table [x=Step, y=Value, col sep=comma] {memoryefficient.csv};
    \addlegendentry{Standard attn}
    \addlegendentry{Memory-efficient attn}
    \end{axis}
    \end{tikzpicture}
    \caption{BLEU scores of a two Transformer models trained with standard attention and memory-efficient attention.}
    \label{fig:blueplot}
\end{figure}

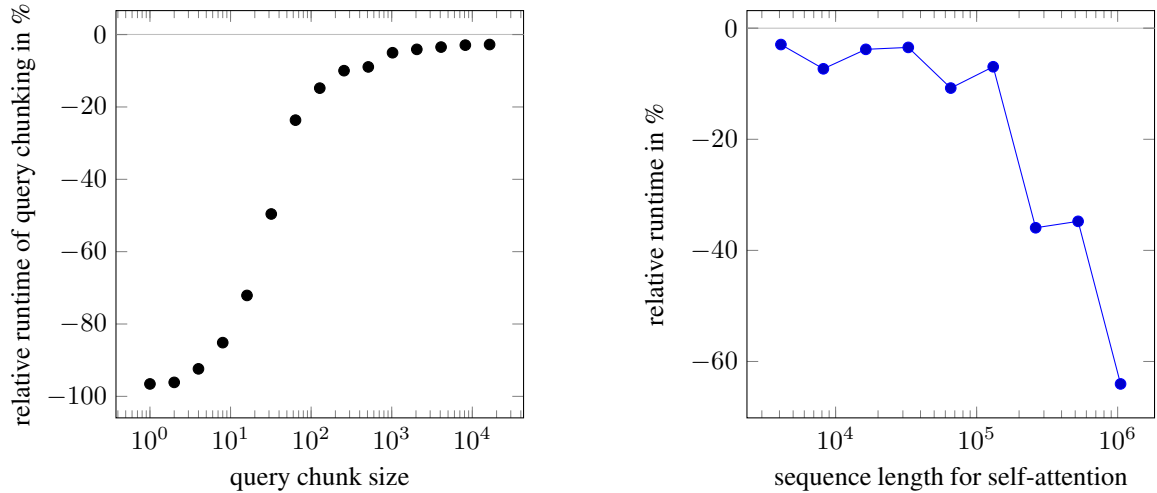
\begin{figure}
    \centering
    \begin{tikzpicture}
    \begin{axis}[ 
    width=7cm,
    height=7cm,
    no marks,
    xmode=log,
    xlabel={query chunk size},
    ylabel={relative runtime of query chunking in \%},
    extra y ticks=0.03949689865112305,
    extra y tick labels=,
    extra y tick style={grid=major},]
    \addplot [only marks] table [x=chunk size, y=runtime, col sep=comma] {query-chunk-runtime.csv};
    \end{axis}
    \end{tikzpicture}
    \qquad\qquad
    \begin{tikzpicture}
    \begin{axis}[
    width=7cm,
    height=7cm,
    xmode=log,
    legend pos=north west,
    xlabel={sequence length for self-attention},
    ylabel={relative runtime in \%},
    extra y ticks=0,
    extra y tick labels=,
    extra y tick style={grid=major},
    ]
    \addplot table [x=input size, y=runtime, col sep=comma] {relative-runtime.csv};
    \end{axis}
    \end{tikzpicture}
    \caption{{\bf Left:} Relative runtime of self-attention on sequence length $2^{15}$ using query chunking compared to standard attention. 
    {\bf Right:} Relative runtime of self-attention using query chunking compared to our memory-efficient algorithm, where both are restricted to the same amount of memory.}
    \label{fig:querychunk}
    \vspace{-3mm}
\end{figure}

We integrated our memory-efficient implementation into a simple Transformer architecture provided in the Flax library, and ran the WMT en-de translation experiment with the standard attention module and with the memory-efficient attention module.
Throughout the training, the two implementations behaved almost identically.
After 100K training steps, the evaluation accuracy reached 62.69 for the memory-efficient implementation and 62.59 for the standard implementation.
This demonstrates that our memory-efficient implementation of self-attention can be used to replace existing implementations.
Figure~\ref{fig:blueplot} illustrates that both models resulted in very similar BLEU scores.
We used the default settings for the WMT en-de experiment as given in the Flax implementation, except that we had to deactivate example packing to simplify the masking code.
This also required us to lower the learning rate to 0.005.

\subsection{Comparison to Query Chunking}
The algorithms introduced in this work chunk both the keys and the queries.
Chunking the only queries has been explored already by~\citet{kitaev2020reformer}, but it is folklore that it slows down the computation significantly.
In Figure~\ref{fig:querychunk} (left), we plot the runtime of self-attention using query-chunking for different query chunk sizes compared to dense self-attention: we see that for small chunk sizes (e.g. $\leq 64$) the performance suffers indeed, but for large chunk sizes, the loss of performance is less significant.
So, while lower memory consumption can be achieved by query chunking alone, small values for query chunking are impractical.

In comparison to query chunking, memory-efficient attention can save additional memory by chunking also the keys. This can help to keep the query chunk size at a desirable point given a fixed memory limit.
In Figure~\ref{fig:querychunk}, we constrained query chunking to the amount of memory that is used by memory-efficient attention with the default settings for key and query chunk size (see Table~\ref{tab:performance}, ``Memory overhead of memory-efficient att.'', we rounded the query chunk size towards the benefit of query chunking).
We see that as the sequence length increases, query chunking eventually slows down significantly as the query chunk size has to be lowered to $\leq 64$, while memory-efficient attention does not suffer a major slowdown (see Table~\ref{tab:performance}, ``Relative compute speed'').
So, in memory-constrained scenarios, memory-efficient attention can outperform query chunking.

\section{Related Work}\label{section:relatedWork}
After publishing our initial draft, we were made aware that \citet{HanhwiJJJJ2019MNNFast} already observed that the division of the softmax operation can be delayed until the end of the attention operation (``lazy softmax''), similar to our Equation \eqref{eq:lazySoftmax}.
But their paper does not discuss memory complexity at all.
They also do not address numerical stability or backpropagation, and, as far as we know, there is no publicly available implementation of their work.
Instead they use this algorithm to reduce the memory \emph{bandwidth} for inference when sharding key-value pairs across multiple chips.

\citet{TriFu2022FlashAttention} provide a CUDA implementation of 
memory-efficient attention and demonstrate that the reduced memory 
requirements can translate to significant speedups on GPUs.
One reason why we do not observe the same performance gains in this paper 
is that standard self-attention already balances the available FLOPs and 
memory bandwidth of TPUs.

\section{Conclusion}

This paper presents a simple trick to reduce the memory requirement of (self-)attention dramatically, which appears to have been simply overlooked by the community.
We hope that this short paper raises awareness of the fact that attention 
is not intrinsically memory-hungry, which may allow us to revisit some of 
the design choices in popular neural architectures and hardware 
architectures.

\section*{Acknowledgements}
We want to thank Andrew Jaegle for discussions on this paper, and for experimenting with memory-efficient attention in the context of Perceiver~\citep{hawthorne2022perceiverAR}.
We are glad to see that the algorithm proposed here has already found interest and would like to thank \citet{rezaei2021github} and \citet{wang2022meap} for reimplementations in JAX and PyTorch with additional features like masking.
We also want to thank DeLesley Hutchins for detailed feedback on our draft.


\bibliography{main}
\bibliographystyle{main}

\end{document}